\documentclass[10pt, a4paper]{article}
\usepackage{lrec2022} 
\usepackage{multibib}
\newcites{languageresource}{Language Resources}
\usepackage{graphicx}
\usepackage{tabularx}
\usepackage{soul}
\usepackage{titlesec}
\titleformat{\section}{\normalfont\large\bfseries\center}{\thesection.}{1em}{}
\titleformat{\subsection}{\normalfont\SmallTitleFont\bfseries\raggedright}{\thesubsection.}{1em}{}
\titleformat{\subsubsection}{\normalfont\normalsize\bfseries\raggedright}{\thesubsubsection.}{1em}{}
\renewcommand\thesection{\arabic{section}}
\renewcommand\thesubsection{\thesection.\arabic{subsection}}
\renewcommand\thesubsubsection{\thesubsection.\arabic{subsubsection}}

\usepackage{epstopdf}
\usepackage[utf8]{inputenc}

\usepackage{hyperref}
\usepackage{xstring}

\usepackage{color}

\title{A Collection of Pragmatic-Similarity Judgments over Spoken Dialog Utterances\\ \vspace*{.5\baselineskip}}

\name{Nigel G. Ward, Divette Marco}

\address{Computer Science, University of Texas at El Paso \\
         El Paso, Texas, USA  \\
        nigelward@acm.org, divettemarco@outlook.com}

\abstract{
Automatic measures of similarity between utterances are
invaluable for training speech synthesizers, evaluating machine
translation, and assessing learner productions. While there exist
measures for semantic similarity and prosodic similarity, there are as
yet none for pragmatic similarity.  To enable the training of such
measures, we developed the first collection of human judgments of
pragmatic similarity between utterance pairs. Each pair 
consisting of an utterance extracted from a
recorded dialog and a re-enactment of that utterance.  
Re-enactments were done under various conditions
designed to create a variety of degrees of  similarity.
 Each pair was rated on a continuous scale by 6 to 9 judges.  
The average inter-judge correlation was as high as 0.72 for English and 0.66 for Spanish.  We make this data
available at
https://github.com/divettemarco/PragSim .
 \\ \newline }

\begin{document}

\maketitleabstract

\section{Motivation}

From a cognitive science perspective, ``similarity is one of the most important relations humans perceive" \cite{richie-bhatia}, as it underlies many aspects of learning, classification, and generalization. From a computational linguistics perspective,  similarity models are important for many applications and 
motivate many representations.  However, there are as yet no models
of {\it pragmatic} similarity.  

Pragmatics includes all the aspects of language use in which people convey information beyond the semantic content.  Pragmatics is especially important in dialog and embodied interaction \cite{marge-espy-wilson-ward-csl}, where people may coordinate action, convey intentions and attitudes, mark topic shifts and connections, manage turn-taking, and so on.   
Measures of lexical and semantic similarity are inadequate for such pragmatic dimensions. 
 This is especially evident for spoken dialog.  For example, two instances of the same lexical content, such as the word {\it okay}, may mean entirely different things, depending on the prosody.  
Conversely, lexically different utterances, such as {\it that's really interesting} and {\it that reminds me of a story} may be functionally very similar, if the prosody
of both politely conveys the intent to close one topic and move on to another. 

Development and evaluation of  models of pragmatic similarity requires a reference set of human
judgments, but such a resource has been lacking. This paper presents the first collection of such judgments.

\section{Application Needs}

Pragmatics is becoming more important for computational purposes, as applications increasingly target natural spoken dialog. This section 
overviews  the needs in two  applications areas.  

Today speech synthesizer output tends to be prosodically neutral and pragmatically uninformative.  
Traditionally speech synthesizer output was evaluated on intelligibility and naturalness, and occasionally also expressivity, but recently there is more interest in evaluation in terms of appropriateness, in part to better support synthesis for dialog applications \cite{omahony21_ssw,wagner-beskow}.
One particular area of interest is synthesis for machine translation, 
where support for conversational uses will need  elements of the
source-language pragmatics to be re-created
in the target-language output
\cite{huang2023holistic,audiopalm,avila-ward23,seamless2023}. 
Evaluation of systems' ability to do so will require a good pragmatic-similarity metric,
for example for evaluating the match between system
output and human-generated reference translations.

Another application area is assessment of human speech and dialog behavior, both for people learning a new language, 
and for people  seeking to overcome speech pathologies.
Existing assessment methods focus on  
evaluating the phonetic, lexical, syntactic or semantic aspects
of production,  but
for communicative effectiveness and social inclusion, it is often the pragmatic aspects of language behavior that matter more. 
We imagine a system which compares a subject's dialog behavior sample to that of an exemplar speaker.
If the behavior of the subject is pragmatically dissimilar to that of the reference speaker
(or any reference speaker in contexts that were pragmatically similar in a corpus), 
then correction or intervention may be appropriate.  
This will, again, 
require a good pragmatic-similarity metric.

Pragmatic-similarity evaluations
are today sometimes proxied by simple prosodic similarity metrics.  
Among other problems, these metrics so far require that the utterances
being compared have the same word sequences.  
However, for both applications, we would ideally like to be able
to judge pragmatic similarity independently of lexical similarity,
in case the speaker or the system chooses to express the same
idea and intent with different words.

\section{Modeling Similarity: Related Work}

Only one previous study seems to have directly targetting pragmatic
similarity \cite{pragst2022}.  This was motivated by the goal of
enabling increased variety of expression for dialog systems.
Unfortunately the generality of the work may be limited, as it was
evaluated only on artificial data and only in terms of
determining whether two sentences embody the same speech act.
Further, the model developed was being purely text-based.

The rest of this section surveys work on two related types of similarity,
semantic and prosodic.   
For convenience, we organize the discussion around 
three approaches for developing similarity models: 
using knowledge, using distributional properties, and using explicit judgments \cite{mihalcea-corley06,chandrasekaran2021evolution}.

{\bf Knowledge-based models}, designed using scientific or common-sense knowledge about what matters to people,   
are prevalent for prosodic similarity.
These are relevant since, as already suggested,
pragmatic meanings are often conveyed by prosody.  Most models of prosodic similarity    focus on intonation (F$_0$) alone
\cite{kominek08,salesky21,hermes98,reichel-similarity,nocaudie16},
with more recent work adding duration features
\cite{mixdorff12,huang2023holistic}, and sometimes others
\cite{rilliard11,me-prosody-ir}.  
However none of these consider all important aspects of prosody, which include 
intensity, timing, rate, phonetic reduction, and voice quality features.
While these are known to sometimes contribute significantly to the 
expression of pragmatic functions, beyond intonation alone \cite{niebuhr-ward},
the extent has never been quantified, for lack of a metric or set of judgments.  
Other work has developed reasonably comprehensive utterance-level
representations of prosody \cite{gemaps}, 
but mostly to support only supervised learning, 
and none tested from the perspective of placing utterances in a space in which distances have meaning.  
The only exception is a very recent paper \cite{avila-ward23}, 
which explored a Euclidean distance metric over 100 features designed to capture 
the prosodic indications of pragmatic functions. 
The quality of this metric was evaluated only on 
post-hoc judgments of its ability to separate  
very similar utterance pairs from very dissimilar ones.  

Another type of knowledge that might be useful for building pragmatic-similarity models is
knowledge of the likely components of pragmatic-similarity perceptions. 
There are a few taxonomies of functions that are important in dialog
\cite{bunt2023semantic,seebauer2023re}, which, although clearly 
incomplete, can guide us in useful directions.  
It does seem that any single specific dimensions of pragmatic or stylistic similarity can be handled well, 
if there is data suitable for supervised training.  
This has been done, for example, 
done for judging the  clarity of turn-hold/yield intentions
\cite{ekstedt2023}. However the number of important pragmatic
functions is very large, and it is not clear how far such
taxonomy-based models could advance us to a general model of pragmatic similarity.  

{\bf Distribution-based models} exploit the general association
between  occurring in similar contexts and being perceived similarly.
Self-supervised learning uses distributional properties to learn embedding spaces in which any input can
be concisely represented, and proximity in such a space can 
be  a proxy for similarity.  
For semantic similarity, such metrics have
been very useful, and even more so when combined with
knowledge-based algorithms, for example to deal with reorderings \cite{moverscore}.

Although self-supervised learning of pragmatics-capturing representations would seem a natural next step \cite{purver-sadrzadeh}, and self-supervised models of semantic similarity in conversation and of dialog processes
have been created  \cite{yinfei-yang,nguyen2023generative}, 
no general representations of pragmatic information appear to
yet exist.  While some dimensions of pragmatic function are fortuitously captured in speech pretrained models such as HuBert and wav2vec2.0 
\cite{guan-ting}, it is not known how much pragmatic information these models represent,
nor whether this is sufficient for modeling pragmatic similarity.  

{\bf Models trained on human judgments} have been useful
in particular for estimating semantic similarity for machine translation, 
not only for text but also for speech \cite{wieting2019beyond,gehrmann2023repairing,besacier2022textless,chen2022blaser,seamless2023}.  
Some of this work starts with pretrained models, 
greatly reducing
the amount of human-judgment data needed for the final training. 
While not directly relevant to similarity, recent work training models
to predict human judgments of speech-synthesis quality
\cite{voicemos22,speechlmscore,cooper2022generalization},
and of utterance suitability for a given 
dialog context \cite{wallbridge-eacl23}, illustrates what can be done
when appropriate data is available. 

Overall, while it is likely that such  models have some utility also for estimating pragmatic 
similarity, determining how much will require a set of pragmatic-similarity judgments.
In general, whether developing pragmatic-similarity models from scratch or adapting existing models, the field will need a collection of human perceptions of pragmatic similarity.  
We here present the first. 

\section{Pilot Studies}

A good procedure for obtaining judgments should: cover a wide range, be reliable, obtain rich information, be efficient, and be not onerous for judges. We did three pilot studies, with a total of 16 judges, to investigate some options. 

1) For obtaining judgments, we tested three methods: rating, ABX, and odd-one-out. While the latter two 
have their advantages --- being easier for the judges, seemingly having higher agreement, 
and potentially sidestepping some weaknesses of pairwise comparisons \cite{aldrich2009determines}, 
we chose rating of utterance pairs, because it  gives  more information for training models,
and because we thought that perceptive, well-trained judges could handle it.

2) We tried different rating scales. Some judges felt that five options (1 -- 5) were not enough as they wanted more precision, but others felt that even 5 options were too many. Again prioritizing the amount of information 
to obtain and assuming great judges, we chose a continuous scale.

3) When we inadvertently had a delay between the play-back of the utterances
being compared, judging the similarity was felt harder, 
doubtless because of the limited size of auditory memory. 
We decided to play both members of each pair back-to-back, 
separated only by a short beep.

4) We briefly considered presenting the stimuli with contexts, 
rather than as isolated utterances \cite{wagner-beskow}.
This would likely boost the inter-annotator agreement, but would be
much more time-consuming, and the resulting judgments would be harder
to build models for.  

5) We started with stimuli that were random pairs of utterances from a large corpus. Judges noted that the task was strange because the pairs were often extremely different, for example, of widely different durations. 
With such stimuli, it seemed that we would obtain mostly only ``very dissimilar'' judgments. 
We therefore decided to assemble stimuli such that both members of each pair shared a lot, 
as described in the next section.

6) We found that judges generally tended to be close in their
judgments, but with many exceptions. In such cases, we had them
discuss the factors that were affecting their judgments. These often
varied. For example, one judge thought his judgments were highly
affected by the rate of speech, and another judge thought that her
judgments were highly affected by the ending of the utterances. At a
deeper level, some reported being affected by tone or feeling, and
others by confidence, intention, and level of perceived fluency,
spontaneity, energy, or politeness. It would be interesting to compile
a list of the factors affecting these judgments, and to encourage
judges to consider them all.  Instead, we decided to simply accept
that people differ in how they included and weighed the possible
factors underlying these judgments, and not  try to constrain
them in some way to increase agreement scores.

7) For one utterance pair, one judge remarked on speaker differences, 
which led to an interesting discussion of whether the
perceived difference was due to a difference of intent or personality.
Some judges thought that one of the speakers likely had a more
dominant speaking style in general, and that this difference could
be factored out and discarded when judging pragmatic similarity with an utterance of a different speaker.  
This raises deep issues, but we decided to simply add an instruction to 
``ignore speaker differences.''

8) While some judges mentioned 
noticing surprisingly subtle differences, such as the fact that some utterances seem to have been said with a smile,
others mentioned that they found the task difficult or lacked confidence in their judgments.  
We decided we needed to be selective in choosing judges.

\section{Stimulus Preparation}

Because most use cases for a similarity metric involve comparison of mostly fairly similar utterances, 
we wanted to boost the representation of such pairs.  
To this end, each stimulus consisted of a seed utterance and a 
re-enactment. 

\subsection{Re-enactment Methods}

To obtain a diversity of similarity levels, re-enactments were created using one of 6 methods:

{\bf Voice} (VO).  The re-enactor listens to the seed, and re-creates  it as closely as  they can.  We expected these to be rated very similar to the seeds, differing only to the extent that voices and vocal abilities differ among speakers. 

{\bf Words+Context} (WC)  The re-enactor sees a transcript of the seed and hears 5--10 seconds of the preceding context in the dialog, or more if they ask.  They then say the words in a way that they think is appropriate for the context.  We expected these to also be rated highly.

{\bf Lexically Distinct} (LD) The re-enactor listens to the seed, then recreates something with the same meaning and feeling, but with different words. We expected these to be fairly high in pragmatic similarity.  As a side note, the re-enactors usually seemed to try to keep the same prosodic form, but the lexical differences  often necessitated re-adjustments. For those wanting to model how prosody conveys pragmatic meanings, 
these pairs could be useful for learning to disentangle this from lexically-governed prosody.  

{\bf Context-Only} (CO) The re-enactor  hears only the context before the seed utterance, and then says something they think would be an appropriate continuation of the conversation.
This method was used to produce interestingly different re-enactments that
could be  similar to the seeds, to the extent that most conversations have a natural flow
suggesting a single likely continuation in terms of pragmatic function, 
albeit one that can be realized in various lexical forms. 

{\bf Words} (WO) The re-enactor sees only the transcript and is asked
only to ``say it as you might say it in a conversation."  We expected
these to generally have only weak pragmatic similarity. 

{\bf Speech Synthesizer} (SS) As in the previous condition, the
  re-enactor has access only to the transcript, but this time the
  re-enactor is a speech synthesizer rather than a human.  We expected
  these to be generally not very pragmatically similar to the seeds,
  as the prosody of synthesized utterances tends to be pragmatically 
  neutral.

\begin{table}
\begin{center}
    \begin{tabular}{lccc}
    & \multicolumn{3}{c}{seed properties} \\
    & audio & words & context \\
    \hline
    Voice        & s & s & - \rule{0ex}{2.6ex}\\
    Words+Context & - & w & s \\
    Lexically Distinct & s & s & - \\
    Context Only      & - & - & s \\
    Words   & - & w & - \\ 
    Synthesized  & - & w & - \\ 
    \end{tabular}
\end{center}
\caption{Summary of the information provided in each re-enactment condition.  w means written, s means spoken.}
\label{tab:six-methods}
\end{table}

Table \ref{tab:six-methods} summarizes.

\subsection{Sources of Seeds}

As pragmatic functions mostly occur in dialog, we chose to take all
seeds from recorded dialogs.  As we envisage training a metric to be
widely useful, we used mostly recordings from the DRAL corpus
\cite{avila-ward23}, an unstructured-conversation corpus with a good
variety of topics and interaction styles.  As the speakers in DRAL
were from the same population as our judges, and many of the topics
are of common interest, we expected this choice to also help with
retaining interest during the long judgment sessions. The English
seeds were taken from the English-original conversations, and
similarly for the Spanish seeds.

To increase diversity, we supplemented these with a few seeds from other corpora.  
For English these were mostly of task-oriented dialogs, including:
a billing support corpus, a persuasive dialogs corpus, a negotiation corpus, a gameplay corpus,
 and a referent-identification and problem-solving corpus 
 \cite{ward-interspeech05,acosta-ward09,ward-avila-alarcon,ward-abu,pardo2019montclair}.
 For Spanish these were from an interview corpus \citelanguageresource{spanish-in-texas}, 
 and a telephone call corpus \citelanguageresource{callhome-spanish}.
We also included a few children's utterances, in English from the Talkbank Providence corpus \citelanguageresource{providence}, 
and in Spanish from PhonBank \citelanguageresource{llinas-ojea}. 
Not wanting to ask adults to re-enact toddlers' utterances, these
seeds were matched with other utterances by the same child. 

Seeds were selected: 1) to be at least 3 words but no longer than 6
seconds. This was to balance the need to include enough information to
provide a basis for judgment, and the need to make the process of
obtaining judgments fast and easy. The average length was 3
seconds. 2) to be mostly clear in intent and meaning when heard in
isolation, free of excessive disfluencies and laughter, and free of
extreme emotion or strong personality.  We avoided these to make
things easier for our re-enactors. 3) to contain no reported speech,
that is, containing no quotes or rephrasings of something said by a
third party. Such utterances have two levels of meaning, and were
often hard to judge. 4) to be diverse, in terms of speakers, topics,
dialog activities, and pragmatic functions. The pragmatic functions of
the utterances chosen included sharing information, making requests,
making decisions, asking questions, giving directions, giving advice,
calming a conversation partner, sharing opinions, expressing empathy,
making observations, making estimates, confirming something,
etc. Those utterances also expressed a range of emotions and cognitive
processes such as understanding, enthusiasm, anticipation,
frustration, confusion, hesitation, agreement, astonishment, humor,
etc. They covered a variety of topics such as media, school, career
building, various types of games, money, politics, charity, grief,
love, business, etc. There was a balance between male and female
utterances.  Diversity was a goal in order to obtain a wider base for
training a metric, and also to keep the judgment task more varied and
thereby more engaging for our judges.

\begin{figure*}[t!]
\begin{quote}
\textit{How pragmatically similar are the two clips, in terms of the overall feeling, tone, and intent.}

\textit{Try to ignore: $\bullet$ speaker differences, $\bullet$
differences in the words said $\bullet$ 
insignificant differences in pitch, rate, pausing, etc.  }

\textit{Maybe consider: 
$\bullet$
Similarity in the contexts where they would likely appear  
$\bullet$
Similarity in how a listener would likely respond 
$\bullet$
Similarity in how the speaker may have felt (confident, positive, offended, enthusiastic, etc.) 
$\bullet$
Similarity in the dialog activity (correcting a misconception, teasing, holding the floor, asking a question, implying something, etc.) }
\end{quote}
\caption{Rating Instructions}
\label{fig:instructions}
\end{figure*}

In all there were 80 seeds for English and 40 for Spanish. 
Since some methods required the re-enactor to know seed properties that other methods required them not to know,
no single person could do them all. 
We accordingly assigned re-enactors with care: 
for any seed, one person would do Methods 5 and then 1 (WO and VO), 
another person would do Methods 2 and then 3 (WC and LO), and a third would do Method 4 (CO).  
There were four re-enactors, the authors and an additional female and male, 
chosen based on our favorable impressions of their vocal control and range. 
For the synthesized utterances we used two of the more conversational voices
from Amazon Polly  \cite{awspolly}, namely the female Salli and the male Matthew, each for half.
In each stimulus the seed preceded the re-enactment.
In total, there were 220 English stimuli in the first judgment session, 238 English stimuli for the second judgment session,
and 235 stimuli for the third, Spanish session.

\section{Obtaining the Judgments}

\subsection{Judges}

We wanted judges who were were sensitive to the nuances of language
and who had the patience to participate in long sessions.
We therefore issued invitations to selected individuals
who had been exceptionally eager and effective participants
in a previous data collection \cite{dral2023}, and/or were members of our research lab. 
Those who accepted all turned out to be friends, relatives, classmates, or potential classmates at leasat one other judge. Most were in their early 20s.  Only those who were fully
bilingual in Spanish were judges in the third session.  
The first author  served as a judge for the first two sessions, and the second author for the third.  Each session was held on a Saturday, and lasted about 4 hours. 
Judges were compensated fairly generously, with snacks, lunch and USD 70.

\subsection{Instrument}

Wanting continuous judgments, we had judges enter 
their ratings using QuestionPro's slider scale tool, 
accessed via each judge's own digital device.
The range was 1 to 5, with ticks for the integer values. 
The anchor text was ``no similarities" for 1 and ``virtually identical" for 5, with no intermediate
labels \cite{zielinski08}.
Inputs were recorded at a granularity of 0.1. 
To clarify their task, judges were given a handout including the instructions in Figure \ref{fig:instructions}. 

\subsection{Procedure}

Judges came to a quiet room and sat around large tables, 
near the speakers.   
After an explanation of our aims and an overview of the
task, they signed consent forms. (The procedure was judged exempt from
review by our institution's human-subjects committee.) Judges
first heard three anchors: stimuli that we had chosen to illustrate the extremes and a central level of similarity.   
To further promote calibration, for the first ten
stimuli, and periodically throughout, we had them compare their
ratings and discuss the factors that had affected their judgments. We
stressed that convergence was not expected, that we welcomed
differences of opinion, and that in the end we would be mostly using
the averages of everyone's judgments. 

Stimuli were grouped by source corpus, and we briefly described each corpus. 
Within each set, the stimuli were presented in random order.

For the first few dozen stimuli we played each stimulus pair three times, or more if requested.  As the judges got more experienced, two times were generally enough.
By the end, the pace had sped up to  better than two judgments per minute.

\subsection{Comments and Observations}

Some judges noted  that some seeds contained a small laugh or moment of laughed speech, whereas some re-enactments of these did not, causing them to lower their ratings somewhat. Some of the re-enactments from the context-only condition were wildly longer than their seeds. This was not intended and may have been jarring to the judges. The judges noted that judging the children's utterances was hard. Not only was the audio quality poor, but small children's intentions are often unclear.
One judge suggested it would be helpful to watch videos to see the body language, and another suggested that having the context would help.
 All judges seemed engaged across the 4 hours, for every session, 
 probably thanks in part to generous breaks.

\section{The Data Collection}

\begin{table}[tbh]
  \begin{center}
    \begin{tabular}{lrrr}
        & \multicolumn{2}{l}{\hspace*{4.8ex}English} & Spanish \\
  Sessions   & 1 \hspace*{.6ex} & 2\hspace*{.8ex} & 3 \hspace*{.8ex} \\
    \hline
    \\[-8pt]
    stimulus pairs  & 220 & \hspace*{4ex}238 & 235\\
    judges & 9 & 9 & 6\\
    judgments & 1980 & 2142 & 1410 \\
    agreement & 0.45 & 0.72 & 0.66
\end{tabular}
\caption{Basic Statistics}
\label{tab:sizes}
\end{center}
\end{table}

In all, we collected 5532 judgments, as seen in Table \ref{tab:sizes}. This data is freely available at https://github.com/divettemarco/PragSim .
The release includes all the stimulus pairs and their components, namely the seeds and the re-enactments, and the provenance and  source tracks for the seeds and the entire 
re-enactment tracks, to enable modelers to do per-speaker normalizations. 

\section{Correlations and Observations}

\begin{figure}
    \centering
    \includegraphics[width=1.1\linewidth]{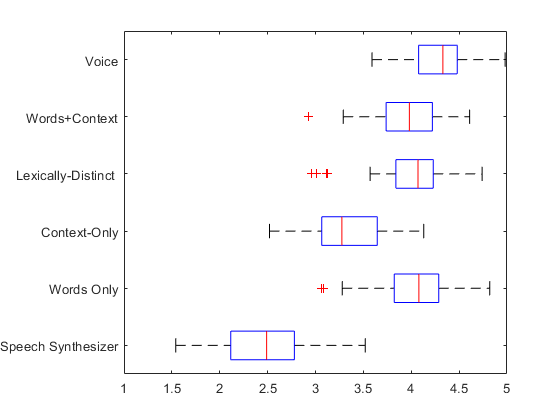}
    \caption{Distributions of the mean per-stimulus ratings for each re-enactment method in Session 1.}
    \label{fig:different-methods-bp}
\end{figure}

Overall there was a good variety of judgments, as seen for Session 1
in Figures \ref{fig:participant-data-entry-bp} and
\ref{fig:different-methods-bp}, with similar patterns seen for the
other sessions. In this respect, our use of diverse seeds and a
variety of re-enactment methods was successful.

The overall inter-judge agreement varied among sessions, but was as high as 0.72, measured as the average of
the pairwise correlations among judges, as seen in Table
\ref{tab:ija}. (We briefly considered using the Spearman
correlation, but scatterplots among pairs of judges did not
reveal any nonlinear patterns.)  

The rest of this section explores the factors affecting
the ratings and the degrees of agreement.

\subsection{ Factors Affecting Ratings}

\begin{figure}
    \centering
    \includegraphics[width=1.1\linewidth]{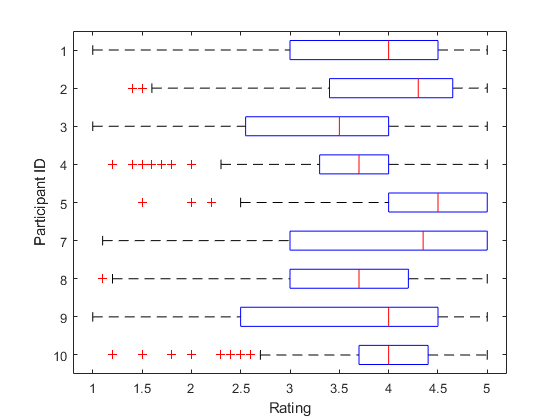}
    \caption{Distributions of judgments for each judge in Session 1.}
    \label{fig:participant-data-entry-bp}
\end{figure}

The judges had different means, some being much more strict, and some using different ranges, as seen for Session 1 in Figure \ref{fig:participant-data-entry-bp}. 
However in Session 2 the judges tended to all use more of the scale, probably reflecting a tendency to calibrate after more experience with the range of stimuli.  

Within each session, judges became slightly more generous over time, with the mean rating per stimulus having a 0.04 correlation with the position in the presentation sequence in the first session, and 0.09 and 0.16 in the second and third.
Perhaps the judges became more familiar
with the re-enactors' voices and better at understanding how their
intents mapped to their speech.

Incidentally, there was a negative correlation between the mean rating and the absolute 
difference in audio duration between the re-enactment and the seed:
$-0.24$, $-.05$, $-0.20$ for sessions 1, 2, and 3, respectively. 
Thus pairs with greater differences in  duration tended to be rated lower.

\begin{table*}[tbph]
  \begin{center}
\begin{tabular}{lrrrrrrrrr}
judge & 1 & 2 & 3 & 4 & 5 & 7 & 8 & 9 & 10 \\
\hline
\\[-8pt]
1 \\
2 & 0.40\\
3 & 0.38 & 0.61\\
4 & 0.37 & 0.59 & 0.59\\
5 & 0.19 & 0.30 & 0.31 & 0.49\\
7 & 0.41 & 0.67 & 0.66 & 0.54 & 0.33\\
8 & 0.39 & 0.64 & 0.60 & 0.80 & 0.40 & 0.54\\
9 & 0.21 & 0.40 & 0.36 & 0.18 & 0.19 & 0.51 & 0.20\\
10 & 0.42 & 0.62 & 0.50 & 0.59 & 0.29 & 0.52 & 0.63 & 0.27\\
 \\[-6pt]
\multicolumn{7}{l}{~~~ Per-Judge Means\rule{2cm}{0ex}} \\
& 0.34 & 0.53 & 0.50 & 0.52 & 0.31 & 0.52 & 0.52 & 0.29 & 0.48\\
\\[-6pt]
\multicolumn{7}{l}{~~~ Overall Mean: 0.45}\rule{2cm}{0ex}
\end{tabular}
\caption{Inter-judge Agreements, Pearson's Correlations}
\label{tab:ija}
\end{center}
\end{table*}

Other trends were apparent on closer examination of the Session 1 data.  As expected, re-enactments generated in the Voice condition were generally rated highly similar to the seeds, and those produced by the Speech Synthesizer only weakly similar. 
Unexpectedly, re-enactments in the Words Only condition were generally more similar to the seeds than those in the Context Only condition. 
We interpret this as indicating that the conversations were less locally deterministic
than we had thought, and that our re-enactors, being familiar with the local population and how they tended to talk about various topics, were often able to infer from the words alone how they were likely to have been said. 
We initially expected re-enactments from the Words+Context condition to be rated 
higher than the Words Only ones, because it gave additional information to the re-enactors. 
However the opposite was true.  
This might be because the limited contexts we provided were more misleading than helpful. 
Although we expected the Lexically Distinct re-enactments to be rated 
higher than those from the Words Only condition, 
the difference was tiny. 
This could be because the judges  focused on the words more than we expected, 
or because re-enactors were able to adequately infer the pragmatics of the seed 
without needing context.

\subsection{Factors Affecting Agreement}


While the level of agreement is probably fully acceptable for many purposes,
we probed the factors behind lower agreement.
As the data was not collected to support a
systematic analysis; the rest of this section only reports the factors
identified using  post-hoc tests.  
Measures for these included inter-judge correlations and the per-item standard deviations,
computed after standardizing (z-normalizing) the ratings of each
judge. 

1) Judge identity was a major factor.
For example, in Session 1 
there was huge variation in
pairwise correlations, ranging from 0.18 to 0.80, as seen in Table
\ref{tab:ija}.  Interestingly, the judge who differed most from the
others (average correlation of 0.29) turned out, on inspection, to
have ratings that were mostly just integers, 
whereas most judges had many  fractional ratings.  In Session 3 we
also noticed that one judge's ratings were mostly very high or very low. 

2) Experience was also a major factor.  
This is seen in the increase in agreement from Session 1 to Session 2, 
as seen in Table \ref{tab:sizes},
although a confounding factor was that the least-agreeing judge did 
not return for that session. 
The effects of experience were also evident within sessions, 
with mostly negative  
correlations between the serial positions of the stimuli and the standard
deviations of the judgments: $-0.13$ and $-0.03$ for Sessions 1 and 2 (but $+0.03$ for Session 3).  
The effects of experience may result in part in convergence in opinions due to the
occasional discussions among judges, and in part from individual judges becoming
more internally consistent in how they used the scale.  

3) There was generally lower agreement for the lower-rated stimuli:
the correlation between the standard deviation and the mean rating was
$-0.60$, $0.00$ and $-.39$ for the three sessions.  This may be because
in part there was a clear upper bound but no obvious reference point
at the lower bound.

4) There was generally lower agreement for the pairs where
the utterances had different lexical content, namely those
using reenactments generated in  the lexically-distinct and context-only 
conditions. Indeed, for Session 2, excluding those stimuli
gave an average inter-judge agreement of 0.80, 
significantly above the 0.72 seen for all stimuli. 
 
5) The specific re-enactment method may have been another factor,
but there were no consistent patterns across the three sessions. 

6) Duration differences between seed and re-enactment were another
factor.  For example, in Session 1 the standard deviations of the
judgements correlated 0.30 with the absolute differences in duration
between the seed and the re-enactment.  For the longer re-enactments,
sometimes the tone varied from the beginning to the end, and different
judges seemed to pay more attention to different parts.


To complement the statistical analyses above, we also examined
 individual stimuli where the agreement was low, specifically the 27
from Session 1 for which the standard deviation exceeded 1.0, looking for
commonalities that might explain why the judgments varied so much.
This enabled us to identify three additional factors: 

\begin{table*}[tbph]
    \small
    \begin{center}
    {\begin{tabular}{lp{15.cm}}
1 & Seed Words: {\it What happened? Today you saw her?}\\
& Re-enactment: {\it So? And why? Did you see her today?}\\
& Both seed and re-enactment are clearly following up with interest
on new information.  Only the seed includes an initial noisy nasal inhalation and had a  wide pitch range, and seems casual, lively, and warm.     (010F\_M3n\_EN\_098r\_17) \\
\\[-8pt]
2 & Seed Words: {\it I get you. I wanna watch it but, it's also really long.}\\
& Re-Enactment: {\it Got it... okay so, they're aliens, um... but like, are they bad or... are they just minding their own business?}\\
& Both seed and re-enactment express understanding and the wish to contribute to the topic without having much to say.  The seed clearly closes the topic while the re-enactment aims to continue the topic.  (035F\_M4l\_EN\_043r\_12)\\
\\[-8pt]
3 & Seed Words: {\it But I- I'm looking forward to u-uh, impressing them and giving them a fun show.}\\
& Re-enactment: {\it We can definitely help you with that.}\\
& Both seed and re-enactment are positive and dominant in tone.
The seed is more creaky and less fluent, and sounds more sincere and engaged. (329F\_M4n\_DC\_1i\_02) \\
\\[-8pt]
4  &  Seed and Re-enactment Words: {\it I was looking through the gifts and... it was for my mom}\\
& Seed and re-enactment are lexically identical, but only the seed includes laughed speech, on the last four words, and overall conveys the intent to tell  a funny story. (019F\_M2l\_EN\_034l\_2)\\
\\[-8pt]
5  & Seed Words: {\it And in that space there’s a monastery.}\\
& Re-enactment: {\it You keep going straight for a couple of streets and that's how you get there.}\\
& Both seed and re-enactment are providing information and have a helpful, confident tone,
but the seed suggests continuation while the re-enactment prosody implies
a  turn end.  (389F\_M4l\_MMTg\_01)
\\
\\[-8pt]
6  & Seed and Re-enactment Words: {\it Wow that’s so humble of you, I never would’ve guessed.}\\
& Seed and re-enactment are lexically identical, but the seed is teasing and clearly sarcastic, while the synthesized re-enactment is  prosodically neutral, although this could be interpreted as deadpan sarcasm. (123F\_M6s\_EN\_081r\_26)\\
\end{tabular}}
\caption{Examples of  Stimuli with Poor  Agreement}
\label{tab:lowest-agreement}
\end{center}
\end{table*}

7) Many of these were for re-enactments using
the Context-Only method.  Sometimes the semantics was wildly different
from those of the seed, as seen in Examples 2, 3, and 5 of Table
\ref{tab:lowest-agreement}.  Perhaps some judges were able to look
beyond this and focus on the pragmatic similarities, while others found
this harder. This is not surprising: we always expected that training
a model to work well for lexically different content will be harder;
it seems harder for people too.

8) Some of these had seeds with unusual vocal properties, 
including ingressive fillers, laughter, falsetto, and strong
breathiness, which were often not present in the re-enactments, as seen in
Examples 1 and 4 of Table \ref{tab:lowest-agreement}.  Perhaps some
judges were more critical of such differences whereas others perhaps
were forgiving.  In particular, it was possible that some judges
interpreted ``ignore speaker differences'' as implying the need to not
penalize re-enactments made by re-enactors with more limited vocal
range (often males).

9) Some of these may have involved  personality perception differences, 
as some of of the seeds conveyed an unusual
degree of warmth, engagement, or scintillating personality, as suggested by 
Examples 1, 3, and 4 of Table \ref{tab:lowest-agreement}.  Again, some
judges may have been less swayed by such differences, and able to
forgive the omission of properties incompatible with
the re-enactor's personality or expressive abilities.

\subsection{Implications}

For those seeking to use this data to build models for automatically estimating pragmatic similarity judgments,
we generally recommend using the Session 1 data for training
and the Session 2 data for evaluation. 
In addition, to avoid dominance by the wider-range judges, it may be better to use as targets, rather than the simple
average, the results of averaging after z-normalizing
each judge's ratings (and optionally then rescaling to 1--5 if desired). 

For some purposes, modelers may wish to consider
excluding stimuli that are difficult to model or are irrelevant to the specific intended use.  These may include stimuli that are not lexically equivalent, differ too much in duration,  
have unusual vocal properties, 
or have a gender difference between seed speaker and re-enactment speaker.

For those wishing to use our methods to collect new sets of similarity judgments, 
we recommend the careful screening,
training, and evaluation of judges. 
It might also be appropriate for some purposes to exclude some of the stimulus types mentioned 
in the previous paragraph, to modify the Context-Only method to include informing the 
re-enactors of the approximate duration of the seed utterances, and 
to provide judges with more specific instructions or illustrations illustrating what aspects of speaker differences to ignore.

\section{Summary and Prospects} 

1) We contribute a carefully-designed and tested protocol.  We hope
others will use this, for example to collect similarity judgments for
other languages.  

2) We also contribute observations on the factors
affecting ratings and agreements, which we hope will inform future
data collection and modeling efforts.

3) Our primary contribution is the ratings.  
These will enable the training of models to approximate
perceived pragmatic similarity.  These models will serve as general
infrastructure, supporting the development of more
conversationally-adept speech-to-speech translation systems, more
effective dialog systems, more precise assessment of communication
skills, and other applications.

\subsection*{Acknowledgements}

We thank Jonathan Avila, Dimitri Lyon, Lidia Almendariz and Raul Gomez for discussions and re-enactments, and Andres Segura for help with the analysis. 
This work was supported in part by the AI Research Institutes program of the National Science Foundation and the Institute of Education Sciences, U.S. Department of Education through Award \# 2229873 -- National AI Institute for Exceptional Education.

\section{Bibliographical References}\label{reference}

\bibliographystyle{lrec2022-bib}
\bibliography{bib}

\section{Language Resource References}
\label{lr:ref}
\bibliographystylelanguageresource{bib}
\bibliographylanguageresource{languageresource}

\end{document}